\documentclass[a4paper]{article}

\usepackage{INTERSPEECH2022}
\usepackage{booktabs}
\usepackage{multirow}
\makeatletter
\renewcommand{\maketag@@@}[1]{\hbox{\m@th\normalsize\normalfont#1}}%
\makeatother

\title{Calibrate and Refine! A Novel and Agile Framework for ASR-error Robust Intent Detection}
\name{Author Name$^1$, Co-author Name$^2$}
\name{Peilin Zhou$^{1,*}\thanks{$^{*}$ Corresponding Author.}$, Dading Chong$^{2}$, Helin Wang$^{2}$,Qingcheng Zeng$^{1}$}
\address{
$^{1}$Zhejiang University, Hangzhou, China\\
$^{2}$Peking University, Shenzhen, China
}
\email{zhoupalin@gmail.com,1601213984@pku.edu.cn,wanghl15@pku.edu.cn,qingchengzeng@outlook.com}

\begin{document}

\maketitle
\begin{abstract}
The past ten years have witnessed the rapid development of text-based intent detection, whose benchmark performances have already been taken to a remarkable level by deep learning techniques.
However, automatic speech recognition (ASR) errors are inevitable in real-world applications due to the environment noise, unique speech patterns and etc, leading to sharp performance drop in state-of-the-art text-based intent detection models. 
Essentially, this phenomenon is caused by the semantic drift brought by ASR errors and most existing works tend to focus on designing new model structures to reduce its impact, which is at the expense of versatility and flexibility. 
Different from previous one-piece model, in this paper, we propose a novel and agile framework called CR-ID for ASR error robust intent detection with two plug-and-play modules, namely semantic drift calibration module (SDCM) and phonemic refinement module (PRM), which are both model-agnostic and thus could be easily integrated to any existing intent detection models without modifying their structures.
Experimental results on SNIPS dataset show that, our proposed CR-ID framework achieves competitive performance and outperform all the baseline methods on ASR outputs, which verifies that CR-ID can effectively alleviate the semantic drift caused by ASR errors.
\end{abstract}

\section{Introduction}
\begin{figure}[t]
  \centering
  \includegraphics[width=75mm]{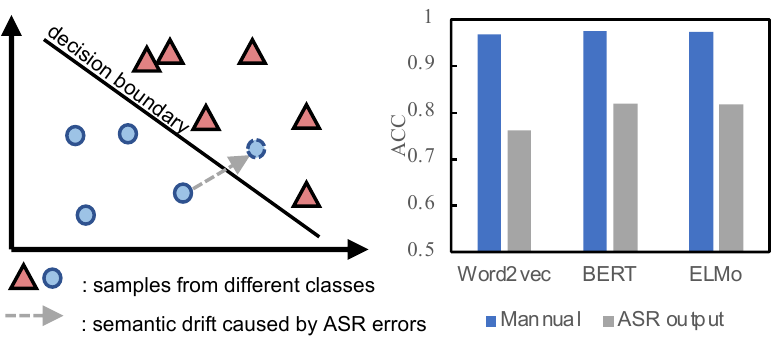}
  \caption{Semantic drift problem (Left) and the comparison of different ID models' performance on manual transcriptions and ASR outputs (Right)}
  \label{fig:intro}
\end{figure}

Intent detection (ID), as one of the key tasks in spoken language understanding, aims to identify users' intents from their utterances.
Driven by advances in deep learning technology, the ID research has entered into a stage of rapid development.
Specifically, many classical methods like convolutional neural network (CNN)  \cite{TurDHH12,XuS13,ZhangFDY16}, recurrent neural network (RNN) \cite{RavuriS15,LiuL16,WangSJ18}, graph neural network (GNN)  \cite{HuWLSC09} and self-attention mechanism \cite{saintent,mengyang} have been explored for this task and obtained superb performance on benchmark datasets.
Moreover, pre-trained language models \cite{qianchen} have also been utilized to better understand the meaning of the user sentences and thus could help to classify the intents more accurately.
Notwithstanding the favorable results of these models, they often assume that automatic speech recognition (ASR) never makes any mistakes.
The training and testing of these ID models are both conducted on error-free manual transcriptions, rather than ASR outputs.
Unfortunately, this overly idealistic setting would make it hard to deploy existing ID models in real-world applications, where ASR errors are unavoidable due to the complex conditions like environment noise and diverse speaking styles or accents.
As shown in the right side of Figure~\ref{fig:intro}, although pretrained language models (LMs) like BERT \cite{DevlinCLT19} and ELMo \cite{elmo} could provide more robust representations compared with static embeddings like Word2Vec \cite{w2v}, they still suffer from sharp performance drop when tested on ASR outputs.
This is because the original representations of user utterances are prone to be distorted by ASR errors (as shown in the left side of Figure~\ref{fig:intro}), which is named as semantic drift problem in this paper.
%

%
Recently, several studies were introduced to mitigate such semantic drift problem.
%
\cite{ChenPB18,HaghaniNBCGMPQW18,LugoschRITB19} proposed to remove the ASR component and extract semantics directly from the speech signals in an end-to-end manner.
Following this trend, \cite{WangWCXN20} applied the mask strategy to audio frames and utilized large-scale unsupervised pre-training technique to learn acoustic representations for SLU.
However, compared with pipeline-based methods, these end-to-end models are less interpretable and more data-hungry. 
In addition, the annotation process of audio data is usually both expensive and time-consuming, which is impractical for industrial applications.
%
Therefore, some researchers proposed to utilize text and speech together for SLU systems. 
For example, \cite{KimKL021} proposed a novel ST-BERT and designed two new cross-modal language modeling tasks to better learn the semantic representations of speech and text modalities.  
\cite{ChungZZ21} suggested to carry out both speech and language understanding tasks during pre-training and constructed a novel speech-language joint pre-training framework for SLU. 
Though achieving excellent performance, they still require pre-training with large-scale datasets, which are not available in some data scarce domains. 
Another branch of ASR-error robust research is to reduce the impact of semantic drift by considering the acoustic similarity between words \cite{HuangC20} or directly injecting phoneme information to the modeling process \cite{0003WZ21}, which had a similar motivation with ours.
But most of them only focused on designing new model structures for specific scenario, and usually show poor compatibility with other methods. So far, designing a both versatile and flexible model has still not been well explored in this research field.
\begin{figure*}[t]
  \centering
  \includegraphics[width=\textwidth]{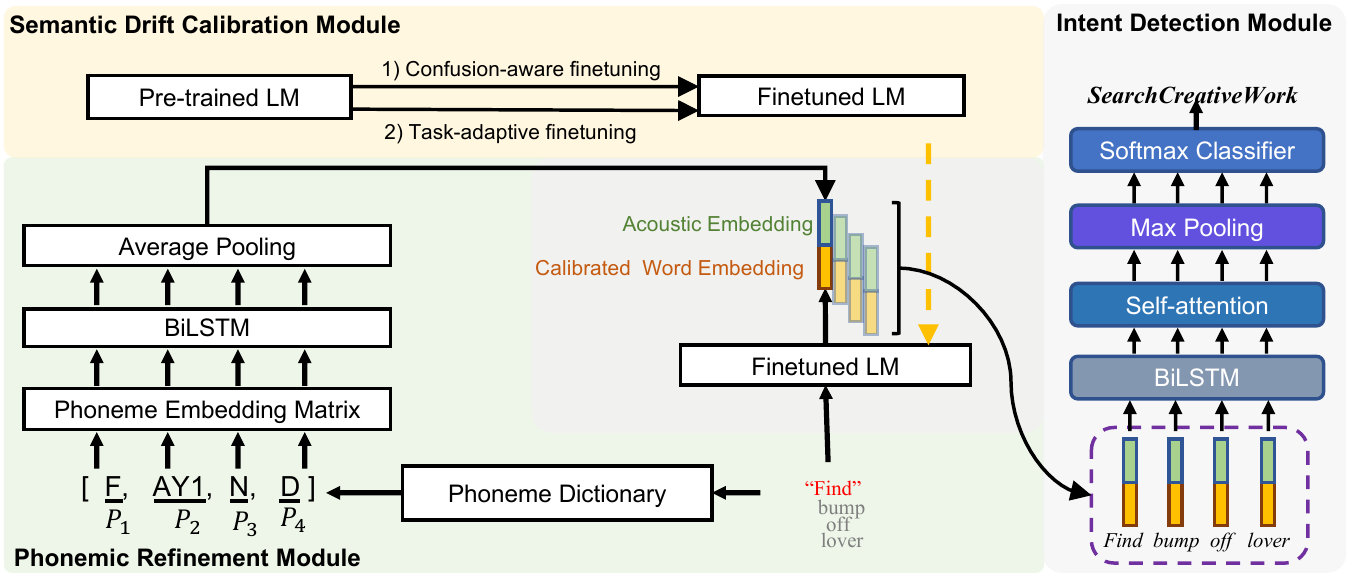}
  \caption{Overview of the proposed CR-ID framework.}
\label{fig:overview}
\end{figure*}

To overcome above-mentioned limitations, we propose a novel and agile framework called Calibration and Refinement for Intent Detection (CR-ID).
Different from previous solutions, our approach decouples the semantic calibration and intent classification process, thus any existing text-based intent detection models could be incorporated into this framework and become more robust to ASR errors.
Specifically, we design two plug-and-play modules to calibrate the semantic drift and refine the calibrated representation with phonemic information, which provides useful signals for the intent classification process.
Our proposed framework will be further detailed in section \ref{sec2} and our main contributions could be summarized as follows:
\begin{itemize}
    \item We propose the CR-ID framework, which could effectively reduce the impact of semantic drift on existing text-based intent detection models without any structural modifications.
    \item We design two plug-and-play modules, namely SDCM and PRM, to calibrate both word-level and sentence-level representation for ASR outputs and utilize the phonemic information to refine and enrich the calibrated representations.
    \item We conduct comprehensive experiments on SNIPS dataset and the results show that, compared with the best baseline model, the intent accuracy and Macro-F1 score of our proposed CR-ID are increased by 1.99\% and 1.86\% respectively, which demonstrates the effectiveness of CR-ID on boosting the robustness of existing ID model.

\end{itemize}

\section{The Proposed Approach}
\label{sec2}
In this section, we present our CR-ID, which is able to effectively and flexibly alleviate the semantic drift problem without changing the structure of classical text-based ID models. The overall architecture of CR-ID is illustrated in Figure\ref{fig:overview}, consisting of three main modules: \textit{Semantic Drift Calibration Module} (Sec. \ref{2.1}), \textit{Phonemic Refinement Module} (Sec. \ref{2.2}), and \textit{Intent Detection Module} (Sec. \ref{2.3}).
\subsection{Semantic Drift Calibration Module}
\label{2.1}
SDCM aims to calibrate the distorted representations of ASR outputs and minimize the negative impacts brought by semantic drift.
To achieve this, inspired by the great success of pretrained language models (PLM) and finetuning techniques, we propose to adopt two PLM finetuning strategies, namely confusion-aware finetuning and task-adaptive finetuning, which are transformed from \cite{HuangC20} and \cite{RuderH18}.

For confusion-aware finetuning, we first use both minimum edit-distance (MED) and word confusion network (WCN) to extract acoustic confusion, which are introduced by \cite{HuangC20}. Due to space limitation, readers could check the details from their paper. 
Taking the two different utterances $x_{1}$ and $x_{2}$ as an example, we use $C=\left\{c_{1}, c_{2}, \cdots, c_{|C|}\right\}$ to denote the set of all acoustic confusions, where $c=\left\{w_{t_{1}}^{x_{1}}, w_{t_{2}}^{x_{2}}\right\}$ consists of two acoustically similar words  $w_{t_{1}}^{x_{1}}$ and $w_{t_{2}}^{x_{2}}$. 
Finally, we propose a new confusion loss to minimize the mean square error (MSE) between the word-level representations and sentence-level representations generated by pretrained language model as follows:
\begin{small}
\begin{equation}
\label{equ1}
\begin{aligned}
    \mathrm{L}_{\mathrm{ca}}\!=\!\frac{1}{|c|} \sum_{c \in C} \sum_{i=0}^{1} \operatorname{MSE}\left(h_{t_{1}, i}^{x_{1}}, h_{t_{2}, i}^{x_{2}}\right)+\operatorname{MSE}\left(h^{x_{1}}, h^{x_{2}}\right)
\end{aligned}
\end{equation}
\end{small}

Task-adaptive finetuning is a widely used technique especially when domain mismatch problem happens, which could effectively adapt pretrained LM from general corpus to the target data. 
For example, given a pre-trained ELMo model and a sentence  $x =\left\{w_{1}, w_{2}, \ldots, w_{|x|}\right\}$, we could directly use the pretraining loss of ELMo as the task-adaptive loss, which could be written as:
\begin{equation}
\label{equ2}
\begin{aligned}
    \mathrm{L}_{\mathrm{ta}}=\frac{1}{|x|} \sum_{t=1}^{|x|}-\log p\left(w_{t} \mid w_{<t}\right)-\log p\left(w_{t} \mid w_{>t}\right)
\end{aligned}
\end{equation}
where $p\left(w_{t} \mid w_{<t}\right)$ and $p\left(w_{t} \mid w_{>t}\right)$ denote the probabilities of $w_{t}$ calculated from forward and backward directions.

Eventually, we jointly finetune the LM using above-mentioned two strategies in a multi-task learning manner and the final loss is as follows:
\begin{equation}
\label{equ3}
\begin{aligned}
    L=L_{\mathrm{ta}}+\lambda L_{\mathrm{ca}}  
\end{aligned}
\end{equation}
where $\lambda$ represents a balancing hyperparameter to control the contribution of each finetuning strategy.

\subsection{Phonemic Refinement Module}
\label{2.2}
Phoneme is the smallest pronunciation unit in speech and the phoneme sequence of each word can represent its acoustic information to some extent. 
Therefore, we design PRM to refine and enrich the calibrated representation by injecting phonemic information into the modeling process.

Firstly, each word $w_{t}$ in the ASR output $x_{asr}$ will be transformed into a phoneme sequence $P_{t}=\left\{p_{1}, p_{2}, \cdots, p_{N_{w_{t}}}\right\}$ via a grapheme to phoneme (G2P) conversion algorithm \cite{g2pE2019}, which highly depends on the pronunciation dictionary. 
In this paper we adopt CMU pronunciation dictionary \cite{KominekB04a} constructed by Carnegie Mellon University, which includes 39 types of phonemes and covers more than 130,000 words as well as their corresponding pronunciation information. 
Figure \ref{fig:overview} also shows the process of converting the word "find" into a phoneme sequence "F,AY1,N,D".
Note that for vowels like "AY", there is a stress marker behind them indicating which stress types it belongs to.
Generally, "0" represents no stress, "1" and "2" represent primary stress and secondary stress respectively, which could provide fine-grained acoustic information for intent detection process.
Then, each phoneme sequence will be mapped into the embedding space and be further encoded by a BiLSTM layer as follows:
\begin{equation}
\label{equ4}
\begin{aligned}
    H_{w_{t}}=& BiLSTM([e_{p_{1}},e_{p_{2}}, \dots,e_{N_{w_{t}}}]),
\end{aligned}
\end{equation}
where $e_{p_{i}}$ denotes the embedding of the phoneme $p_{i}$ and $H_{w_{t}}$ represents the hidden representation matrix for word $w_{t}$.

In the end, average pooling method is conducted on these hidden representation matrices to obtain the final acoustic embedding of the whole sentence $x_{i}$:
\begin{equation}
\label{equ5}
\begin{aligned}
    H_{x_{asr}}^{acoustic}=& [h_{w_{1}},h_{w_{2}}, \dots,h_{w_{N}}],
\end{aligned}
\end{equation}
\begin{equation}
\label{equ6}
\begin{aligned}
    h_{w_{t}}=& Average(H_{w_{t}}),
\end{aligned}
\end{equation}
\subsection{Intent Detection Module}
\label{2.3}
As shown in Figure \ref{fig:overview}, the intent detection module is decoupled from other modules, making it possible to incorporate any existing text-based intent detection model to our proposed CR-ID framework. %
Therefore, we adopt a self-attentive intent classification model inspired by \cite{saintent} as the ID module.
The input of ID module is the concatenation of the calibrated word embedding generated by SDCM and the acoustic embedding generated by PRM.
BiLSTM is used to model the long-term dependency in the utterance and the self attention mechanism is adopted to capture the key information from the calibrated and refined representations. 
Max pooling is utilized to obtain the final sentence-level representation, which is further fed to a softmax classifier to predict user's intent.
%
%
\begin{table}
\small
\caption{Overall performance on manual and ASR output. Bold scores represent the highest results of all methods. }
\label{tab1}
\resizebox{\linewidth}{!}{
\begin{tabular}{@{}ccccc@{}}
\toprule
\multirow{2}{*}{Model} & \multicolumn{2}{c}{Mannual}     & \multicolumn{2}{c}{ASR output}  \\ \cmidrule(l){2-5} 
                       & ACC\%          & Macro-F1\%     & ACC\%          & Macro-F1\%     \\ \midrule
Random                 & 96.87          & 96.91          & 78.60          & 79.87          \\
GloVe                  & 97.15          & 97.18          & 77.12          & 77.70          \\
Word2Vec               & 96.73          & 96.81          & 76.14          & 77.02          \\
FastText               & 97.01          & 97.00          & 79.15          & 79.48          \\ \midrule
BERT (w/o Fine-tuning)      & 96.43          & 96.44          & 80.40          & 80.81          \\
BERT (w Fine-tuning)     & \textbf{97.59} & \textbf{97.70} & 82.01          & 82.80          \\
ELMo (w/o Fine-tuning)       & 96.70          & 96.77          & 80.60          & 81.61          \\
ELMo (w Fine-tuning)     & 97.28          & 97.31          & 81.69          & 82.24          \\ \midrule
SpokenVec (MED)        & 97.01          & 97.21          & 88.52          & 89.23          \\
SpokenVec (WCN)        & 97.04          & 97.12          & 89.55          & 89.97          \\ 
CR-ID (MED)          & 97.42          & 97.50          & 90.85          & 91.32          \\
CR-ID (WCN)            & 97.14          & 97.23          & \textbf{91.54} & \textbf{91.83} \\ \bottomrule
\end{tabular}%
}
\end{table}
\section{Experiments}
\subsection{Dataset}
\label{3.1}
In \cite{HuangC20}, the authors used three datasets, namely SNIPS, ATIS and Smartlight, for their experiments. However, both ATIS (with confusion words) and Smartlight are not available for public because of the copyright issue. Therefore, for fair comparison with the method proposed in \cite{HuangC20}, we directly use their released version of SNIPS dataset to conduct all the experiments. Different from the original SNIPS dataset, \cite{HuangC20} extracted confusion words via the two strategies introduced in Sec\ref{2.1} and added them to the original dataset, which is convenient for researchers to reproduce their results and make improvements on it. The readers could check the details of this dataset in https://github.com/MiuLab/SpokenVec.

\subsection{Baselines and Implementation Details}

A number of ASR error robust ID models have been proposed in the past few years. We do not compare with all of them because
many previous methods are not directly comparable due to the use of different model architectures.
Hence, we select SpokenVec \cite{HuangC20} and construct several baselines that are fair (use the same information, similar architectures, etc.) to compare with.
Specifically, we use the intent detection module (introduced in Sec \ref{2.3}) as the base model, because the self-attentive intent detection model has already achieved comparative performance on SNIPS dataset according to \cite{QinXC021}. Then we incorporate it with different word embedding techniques as the baselines.
\\
\textbf{Static Word Embedding}. We use three pre-trained static word embeddings, Word2Vec \cite{w2v}, GloVe \cite{PenningtonSM14} and FastText \cite{JoulinGBDJM16}, as the embedding matrix to help encode sentences. We also use a randomly initialized embedding matrix as a comparison.\\
\textbf{Contextual Word Embedding}. We evaluate two pretrained language models, ELMo and BERT, to obtain contextual word embeddings. And each LM is evaluated with fixed and unfixed parameters respectively.\\
\textbf{Implementation Details}. For the ID base model, the  dimension of BiLSTM and self-attention layer are all set to 300, the number of heads is set to 8, the batch size is 64. All ID base models are trained on the manual transcribed training set for 50 epochs using Adam optimizer with learning rate as 3e-4, and then tested on the manual transcriptions and ASR outputs respectively. 
For the SDCM, for fair comparison with SpokenVec, we follow its setting and adopt ELMo as the pretrained LM. We train the SDCM for 10 epochs with the batch size of 32, learn rate of Adam set to 1e-4.
For PRM, the dimension of the embedding and BiLSTM hidden layers are all set to 50 and the PRM is jointly trained with ID base model.

\begin{table}[t]
\centering
\caption{Ablation study}
\label{tab2}
\resizebox{\linewidth}{!}{
\begin{tabular}{@{}ccccc@{}}
\toprule
\multirow{2}{*}{Model}                 & \multicolumn{2}{c}{Mannual}     & \multicolumn{2}{c}{ASR output}  \\ \cmidrule(l){2-5} 
                                       & ACC\%          & Macro-F1\%     & ACC\%          & Macro-F1\%     \\ \midrule
Full                                   & 97.14          & 97.23          & \textbf{91.54} & \textbf{91.83} \\
w/o acoustic embedding                 & 96.71          & 96.81          & 90.55          & 90.87          \\
w/o confusion-aware fintuning strategy & 97.15          & 97.21          & 88.60          & 89.01          \\
w/o task-adaptive finetuning strategy  & \textbf{97.28} & \textbf{97.41} & 82.12          & 83.14          \\ \bottomrule
\end{tabular}%
}
\end{table}
\begin{table}[t]
\centering
\caption{Performance comparison of using different distance functions as the confusion loss.The metric is accuracy.}
\label{tab3}
\resizebox{\linewidth}{!}{
\begin{tabular}{@{}cccccc@{}}
\toprule
\multirow{2}{*}{Test data type} & \multirow{2}{*}{Confusion extraction type} & \multicolumn{4}{c}{Type of loss function} \\ \cmidrule(l){3-6} 
       &     & Cosine         & L1    & MSE            & Triplet \\ \midrule
ASR    & MED & 90.56          & 84.90 & \textbf{90.85} & 88.88   \\
ASR    & WCN & 90.82          & 88.26 & \textbf{91.54} & 90.10   \\
Manual & MED & 96.55          & 92.71 & \textbf{97.42} & 96.40   \\
Manual & WCN & \textbf{97.30} & 96.86 & 97.14          & 96.73   \\ \bottomrule
\end{tabular}%
}
\end{table}
\begin{figure}[t]
    \centering
    \includegraphics[width=6.5cm]{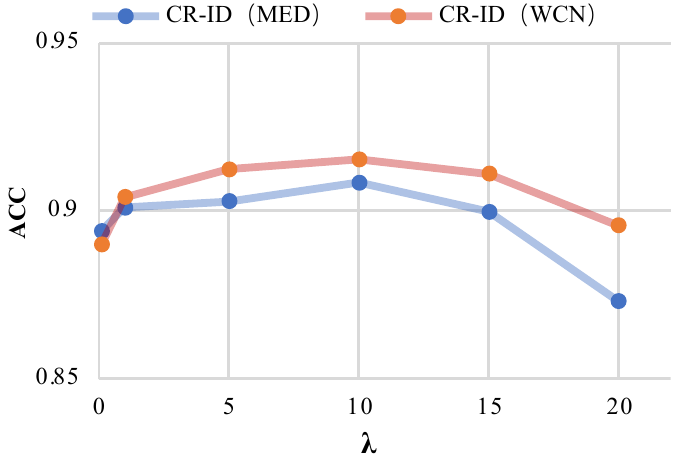}
    \caption{Parameter sensitivity of $\lambda$}
    \label{fig:param-attribute}
\end{figure}
\subsection{Results and Analysis}
The overall performance of the baselines and CR-ID are summarized in Table 1.
Note that as introduced in Sec \ref{2.1} and Sec \ref{3.1}, the confusion word pairs could be generated by minimum edit distance (MED) or word confusion network (WCN). 
Hence, for SpokenVec and our proposed CR-ID, we also report the performance variations using different confusion extraction methods in Table \ref{tab1}.
Here are some observations from the Table \ref{tab1}:
when testing on the manual transcriptions, the performance scores of all methods are very close, and the method based on contextual word embedding is slightly better than the static counterparts.
However, the performances of all baselines except for SpokenVec  drops sharply on the ASR output, demonstrating the necessity to reduce the negative impacts caused by semantic drift problem.
CR-ID (WCN) achieved the best performance in terms of both Accuracy and Macro-F1. 
Specifically, compared with the best static word embedding based baselines, the Accuracy and Macro-F1 of CR-ID (WCN) are increased by 12.39\% and 11.96\% respectively; compared with the best contextual word embedding based baseline model, the performance are improved by 9.5\% and 9\% respectively; even compared with the SpokenVec, which is a very strong baseline, the performance gains still achieve 1.99\% and 1.86\% respectively, demonstrating the effectiveness of our propose CR-ID framwork.
\subsection{Ablation Study}
In order to figure out the contribution of different modules in our proposed CR-ID, we conduct ablation study for each plug-and-play module, as shown in Table.\ref{tab2}:
1) CR-ID w/o acoustic embedding, which only use SDCM for calibration;
2) CR-ID w/o confusion-aware finetuning strategy, where only task-adaptive finetuning and PRM are reserved.
3) CR-ID w/o task-adaptive finetuning strategy, where only confusion-aware finetuning and PRM are reserved. 
We observe that all these components contribute to performance improvements when testing on the ASR outputs. Specifically, task adaptive fine-tuning strategy contributes the most to the  robustness of ID module. 
When this strategy is removed from the CR-ID, the accuracy and Macro-F1 are decreased by 9.42\% and 8.69\% respectively.
And the confusion-aware finetuning strategy take the second place. Without it, the accuracy and Macro-F1 are decreased by 2.94\% and 2.82\% respectively.
Without acoustic embedding, the accuracy and Macro-F1 are decreased by 0.99\% and 0.96\% respectively.
Therefore, the combination of SDCM and PRM could significantly improve the robustness of ID module to ASR errors.

In addition,  we also explore the effect of different distance functions in the confusion loss on model's performance. Here we select three classical distance functions (Equation \ref{equ7},\ref{equ8},\ref{equ9}) to subsitute the MSE in Equation \ref{equ1}, and the results are shown in Table \ref{tab3}. We observe that MSE achieves the best performance under most experimental settings, which is the reason why we finally choose MSE distance for the confusion-aware finetuning strategy. 
\begin{tiny} 
\begin{equation}
\label{equ7}
\begin{aligned}
    \mathrm{L}_{\text{cos}}\!=\!\frac{1}{|C|}\sum_{c \in C}\sum_{i=0}^{1}\left(1-\frac{h_{t_{1}, i}^{x_{1}} \cdot h_{t_{2}, i}^{x_{2}}}{\left\|h _ { t _ { 1 } , i } ^ { x _ { 1 } } \right\|\left\| h_{t_{2}, i}^{x_{2}}\right\|}\right)+\left(1-\frac{h^{x_{1}} \cdot h^{x_{2}}}{\left\|h ^ { x _ { 1 } } \right\|\left\| h^{x_{2}}\right\|}\right)
\end{aligned}
\end{equation}
\end{tiny}
\begin{tiny} 
\begin{equation}
\label{equ8}
\begin{aligned}
    \mathrm{L}_{\mathrm{l1}}=\frac{1}{|C|} \sum_{c \in C} \sum_{i=0}^{1}\left\|h_{t_{1}, i}^{x_{1}}-h_{t_{2}, i}^{x_{2}}\right\|_{1}+\left\|h^{x_{1}}-h^{x_{2}}\right\|_{1}
\end{aligned}
\end{equation}
\end{tiny}
\begin{tiny} 
\begin{equation}
\label{equ9}
\begin{aligned}
\begin{gathered}
    \mathrm{L}_{\!\text{triplet}}\!=\!\frac{1}{|C|} \sum_{c \in C} \sum_{i=0}^{1} \text{triplet}\left(h_{t_{1}, i}^{x_{1}}, h_{t_{2}, i}^{x_{2}}, h_{t_{3}, i}^{x_{3}}\right)\!+\!\text{triplet}\left(\!h^{x_{1}},\!h^{x_{2}},\!h^{x_{3}}\right) \\
    \text{triplet}(a, p, n)=\max \left\{d\left(a_{i}, p_{i}\right)-d\left(a_{i}, n_{i}\right)+\text{margin}, 0\right\} \\
    \quad d\left(x_{i}, y_{i}\right)=\left\|x_{i}-y_{i}\right\|_{p}
\end{gathered}
\end{aligned}
\end{equation}
\end{tiny}
\subsection{Hyperparameter Sensitivity}
In this section, we aim to analyze the effect of the balancing hyperparameter $\lambda$ (in the Equation \ref{equ3}) on the performance of CR-ID. The results are illustrated in Figure \ref{fig:param-attribute}. It can be observed that for both CR-ID (MED) or CR-ID (WCN), when $\lambda$ increases from 0.1 to 10, the model performance is slightly improved, but when the $\lambda$ gets larger (e.g. larger than 15), the performance of the model begins to decline. Therefore, 
for all the CR-ID related experiments, we set $\lambda$ to 10 to better balance the impact of task-adaptive finetuning and confusion-aware finetuning on model optimization.

\section{Conclusion}
In this paper, we propose a novel and agile framework, called CR-ID, for ASR error robust intent detection. 
Two plug-and-play modules, namely SDCM and PRM, are designed to calibrate both word-level and sentence-level representation for ASR outputs and utilize the phonemic information to refine and enrich the calibrated representations. 
Experimental results on SNIPS dataset show that our proposed CR-ID outperform all baseline models on the ASR outputs, demonstrating that our proposed framwork could effectively reduce the impact of semantic drift on existing text-based intent detection models and boost their robustness to ASR errors.
\bibliographystyle{IEEEtran}

\bibliography{mybib}


\end{document}